\documentclass[a4paper,twoside]{article}

\usepackage{epsfig}
\usepackage{subcaption}
\usepackage{calc}
\usepackage{amssymb}
\usepackage{amstext}
\usepackage{amsmath}
\usepackage{amsthm}
\usepackage{multicol}
\usepackage{pslatex}
\usepackage{apalike}

\usepackage{etex,etoolbox}
\usepackage{algorithm}
\usepackage[noend]{algpseudocode}
\usepackage{amssymb, amsmath, amsfonts, bbm, latexsym, graphicx, tabularx, tabulary, upgreek, amsthm, dsfont, booktabs, fancyhdr, mathtools, tikz, epigraph}

\usepackage{SCITEPRESS}     

\makeatletter
\newcommand{\longdash}[1][2em]{%
  \makebox[#1]{$\m@th\smash-\mkern-7mu\cleaders\hbox{$\mkern-2mu\smash-\mkern-2mu$}\hfill\mkern-7mu\smash-$}}
\makeatother
\newcommand{\omitskip}{\kern-\arraycolsep}

\begin{document}

\title{The Labeling Distribution Matrix (LDM):\\A Tool for Estimating Machine Learning Algorithm Capacity}

\author{\authorname{Pedro Sandoval Segura\sup{1,2}, Julius Lauw\sup{1}, Daniel Bashir\sup{1}, Kinjal Shah\sup{1}, Sonia Sehra\sup{1}, \\ Dominique Macias\sup{1} and George Monta\~nez\sup{1}}
\affiliation{\sup{1}AMISTAD Lab, Department of Computer Science, Harvey Mudd College, Claremont, CA, USA}
\affiliation{\sup{2}Department of Computer Science, University of Maryland, College Park, MD, USA}
\email{psando@cs.umd.edu, \{julauw, dbashir, kdshah, ssehra, dmacias, gmontane\}@g.hmc.edu}
}


\abstract{Algorithm performance in supervised learning is a combination of memorization, generalization, and luck. By estimating how much information an algorithm can memorize from a dataset, we can set a lower bound on the amount of performance due to other factors such as generalization and luck. With this goal in mind, we introduce the Labeling Distribution Matrix (LDM) as a tool for estimating the capacity of learning algorithms. The method attempts to characterize the diversity of possible outputs by an algorithm for different training datasets, using this to measure algorithm flexibility and responsiveness to data. We test the method on several supervised learning algorithms, and find that while the results are not conclusive, the LDM does allow us to gain potentially valuable insight into the prediction behavior of algorithms. We also introduce the Label Recorder as an additional tool for estimating algorithm capacity, with more promising initial results.}

\onecolumn \maketitle \normalsize \setcounter{footnote}{0} \vfill

\section{\uppercase{Introduction}}
\label{sec:introduction}


Determining the representational complexity of a learning algorithm is a long-standing problem in machine learning. Well-known methods for doing so include upper bounding an algorithm's model complexity using the VC dimension and measuring an algorithm's ability to fit noise with Rademacher complexity. We let \textit{algorithm capacity} denote the representational complexity of the outputs of a learning algorithm, which is a measure of the algorithm's ability to store information in its trained models that can subsequently be used to make inferences--this stored information might take the form of memorized data points or underlying information about how the data was generated. A method capable of providing an \textit{estimate} of an algorithm's capacity instead of bounding it offers concrete progress towards answering the question of whether a given algorithm will overfit or underfit on a particular dataset.

Building a formal framework for the overfitting and underfitting of machine learning models is of prime importance as researchers desire to tune their models in ways that will lead to greater generalization performance. We want to train our models to capture inherent data relationships and be able to perform accurately over unseen data. It is common belief that machine learning algorithms perform better when their capacity is appropriate for the true complexity of the task, such that the algorithm is able to capture relationships within the provided training data. In other words, underfitting is a result of a model having insufficient capacity, while overfitting is a repercussion of models with excess capacity. In this manuscript, we introduce a proxy for model capacity through empirically observable Labeling Distribution Matrices (LDMs).

\subsection{Existing Characterizations}

To the authors' knowledge, there is no general method to empirically measure algorithm capacity for an arbitrary machine learning method. If such a procedure existed, we would make progress toward determining whether overfitting or underfitting could occur on a dataset, given information on the complexity of said dataset. It is important to note the distinctions of different definitions of model capacity as currently understood.

\textbf{Representational capacity}: specifies the family of functions a learning algorithm can choose from when varying the parameters in order to reduce a training objective.

\textbf{Effective capacity}: specifies that imperfections and assumptions made in the optimization learning algorithm can limit representational capacity.

Our method for measuring the entropy of the labeling distributions produced from a classification model aims to capture representational capacity. We train a given classification model on noisy datasets, then measure how the training affects the distribution of output labels. Training a model on noise and analyzing how that affects the labeling distributions that are produced should serve as a reasonable proxy for a classification model's ability to model arbitrarily complex relationships. 

\subsection{Experimental Setup}
Our goal is to characterize the capacity of a classification algorithm by measuring its expressiveness. In this exploratory work, we will focus our attention on determining the capacity of classification models. The procedure involves analyzing the probability distributions over all possible ways of labeling a holdout set, given an information resource that the model was trained on. Note that we use information resource to mean training data within this context. We want to determine to what extent is an arbitrary machine learning classification model capable of capturing the relationship between features and labels for a given set of information resources. Our hypothesis is that if, on datasets without any inherent relationship between features and labels, the model is able to classify correctly beyond what is expected by random guessing then this suggests the model has the storage capacity to memorize elements of the dataset.

As a concrete example, consider a set of $K$ training datasets, each with $N$ data points. Each data point can be labeled as one of $C$ possible classes. We define the $i^{\text{th}}$ training dataset formally as follows:
    \begin{align*}
        f_{i} &= \{(x_{1}, y_{1}), (x_{2}, y_{2}),...,(x_{N}, y_{N}) \} \\
        \text{where } x_j &= j^{\text{th}} \text{ data point } \\
                      y_j &= \text{ label assigned to data point $x_j$} 
    \end{align*}
for all $1 \leq i \leq K$. Also, let us consider a holdout set $H$ of size $N'$:
$$H = \{z_{1}, z_{2}, ..., z_{N'}\}$$
which we label using a trained classification model $M$. Here, $z_j$ is the $j^{\text{th}}$ data point of the holdout set. The label for the data point $z_j$ of the holdout set is $M(z_j)$.
    
    Based on the above definitions, we see that for any classification model trained on dataset $f_{i}$, there is a total of $C^{N'}$ possible labeling combinations for a given holdout set $H$, where $C$ represents the number of classes available for the given dataset. For example, if the training dataset contains binary labelings (where $y_j \in \{0, 1\}$), then $C = 2$ and there are $2^{N'}$ possible labelings for the holdout set. 
    
    \subsection{Simplex Vectors}\label{ssec:simplexvectors}
    We will be using a probabilistic model to evaluate algorithm expressiveness. The first step is to construct a series of simplex vectors $P_{f_{i}}$, each of which is represented as a vector of probabilities of size $C^{N'}$ for a given training dataset $f_{i}$. The idea is that, for each data point $z_{j}$ in a given holdout set $H$, we can find the probability that the trained classification model $M$ will assign each of the possible $C$ classes. By determining the probability distribution of a classification model's assignments over $C^{N'}$ possible labeling combinations for a given training dataset $f_{i}$, we can gain a better understanding of the capacity of the model based on the nature of the resulting probability distributions. If we see that the model typically assigns probability mass on the same subset of possible labelings regardless of the training data, then we can say that the model is less expressive and has lower capacity. We construct the simplex vector $P_{f_{i}}$ for a given training dataset $f_i$ as follows:
    \begin{enumerate}
        \item Train classification model $M$ on $f_i$.
        \item Label holdout set $H$ using $M$.
        \item For every $j^{\text{th}}$ possible labeling $l$ of the holdout set, compute the probability that $M$ would have assigned $l$. Set this probability as the $j^{\text{th}}$ entry of $P_{f_{i}}$.
        \item Normalize $P_{f_{i}}$. 
    \end{enumerate}
    
    Given that each entry in the simplex vector corresponds to the probability of a particular labeling combination for a given holdout set $H$, the entry is computed by taking the product of the probabilities assigned to each class for the given holdout feature sample. Consider a holdout set $H = \{z_1, z_2, z_3 \}$ of size 3. If a given training dataset has classes $0$, $1$, or $2$, then the probability assigned to the entry corresponding to the labeling $\{z_1: 0, z_2: 1, z_3: 2 \}$ will be the product of the probability of the class $0$ being assigned, the probability of class $1$ being assigned, and the probability of class $2$ being assigned. In other words,
    \begin{align*}
        P(M(z_1) = 0, M(z_2) = 1, M(z_3) = 2 \mid z_1, z_2, z_3) &=\\
        &\hspace{-15em}P(M(z_1) = 0)P(M(z_2) = 1)P(M(z_3) = 2)
    \end{align*}
    where we assume that the probability of a single label is conditionally independent of the probability of any other label, since we are randomly generating training datasets.
    
        \subsection{Dirichlet Characterization}\label{ssec:dirichletsection}
    
    Now that we have a series of simplex vectors $P_{f_i}$ for all $1 \leq i \leq K$, we seek to measure the diversity of these probability distributions. Because the support of a Dirichlet distribution can be viewed as a set of probability distributions, we worked to infer the parameters of the Dirichlet distribution from which the simplex vectors $P_{f_i}$ were drawn. The idea is that we can determine how expressive an algorithm based on the diversity of the probability distributions generated over labelings of a given holdout $H$, for different information resources $f_{i}$. We hypothesize that the greater the expressiveness of an algorithm, the more diverse the probability distributions will be, since such algorithms will not tend to have fixed preference for output responses regardless of the training data. An expressive algorithm tends to be more ``flexible'' such that, given multiple information resources $f_{i}$ to train on and a fixed set of holdout features, it will be more responsive to the differences in the different information resources that it trained on. Thus, the more responsive an algorithm is to changes in training data, the resulting simplex vectors for different datasets should show greater diversity.
    
    Supposing the simplex vectors were drawn independently from a Dirichlet distribution, we could use them to infer a vector of alpha priors $\vec{\alpha}$ corresponding to that Dirichlet distribution. We claim that the entropy of this distribution gives us a means to estimate expressiveness because it tells us how uniformly the inferred Dirichlet distribution assigns probabilities to its support. The more uniform the distribution, the more expressive the algorithm, since the simplex vectors could then be drawn from anywhere on the simplex. Diversity of sampled simplex vectors implies more uniform mass over the support. We compute the entropy of the Dirichlet distribution by computing the entropy of $\vec{\alpha}$, which parameterizes the distribution.
    
    \subsection{Connections to Rademacher Complexity}
    
    One can view this technique, of training an algorithm on noise and analyzing the distribution of possible labelings, as a means of better understanding the space of functions that the classification algorithm is reasoning over. Similarly, in computational learning theory, Rademacher complexity provides us with another way of measuring hypothesis space complexity.

\textbf{Definition: }(Rademacher Complexity) Let $\mathcal{H} \subset \mathcal{F} = \{ f: \mathcal{X} \rightarrow \mathbb{R} \}$ be a class of functions we are exploring defined on domain $X \subset \mathcal{X}$, and $S = \{ x_i \}_{i=1}^{n}$ be set of samples generated by some unknown distribution $\mathcal{D}_{\mathcal{X}}$ on the same domain $\mathcal{X}$. Define $\sigma_{i}$ to be uniform random variable on $\pm 1$, for any $i$. The empirical Rademacher complexity or Rademacher average is defined as follows:


$$\hat{\Re}_{S}(\mathcal{H}) = \mathbb{E}_{\sigma} \Bigg[ \underset{f \in \mathcal{H}}{\text{ sup }} \frac{1}{n} \sum_{i=1}^{n} \sigma_i f(x_i) \Bigg] $$


The supremum measures the maximum correlation between $f(x_i)$ and $\sigma_i$ over all $f \in \mathcal{H}$, for a given set $S$ and Rademacher vector $\sigma$. Because we're taking the expectation over $\sigma$, the empirical Rademacher complexity of $\mathcal{H}$ measures the ability of functions within this space to fit random noise. The Rademacher complexity, therefore, can be thought of as the expected noise fitting ability of $\mathcal{H}$ over all data sets $S$ drawn from $\mathcal{D}_{\mathcal{X}}$ \cite{rademachercomplexity}.

\subsection{Estimating Complexity from the Labeling Distribution Matrix (LDM)}
As described in Section \ref{ssec:dirichletsection}, we take an arbitrary classification model $M$, train it on an information resource (training dataset) $f_i$, and build a probability distribution over $\Omega$, the space of all possible labelings of our holdout set $H$. Because a classification algorithm can only provide us with class probabilities given the features of an example, we build probability distributions of length $|\Omega|$, which are simplex vectors. Each element in a simplex vector represents the probability of encountering a particular combination of labeling. Iterating this process over all $K$ training datasets, we eventually generate a series of simplex vectors $P_{f_1}, P_{f_2}, ..., P_{f_K}$ each of length $|\Omega|$ to characterize the probabilities that $M$ places on each element of $\Omega$. This produces our labeling distribution matrix $L$,

\[
L = \begin{bmatrix}
\mid & \mid & & \mid \\
P_{f_1} & P_{f_2} & \cdots & P_{f_K} \\
\mid & \mid & & \mid \\
\end{bmatrix}
\]

Given these $K$ simplex vectors, whose entries sum to $1$ after normalization described in \ref{ssec:simplexvectors}, we propose a method to measure the capacity of model $M$ that works as follows:
\begin{enumerate}
    \item We suppose each of the column vectors in our matrix is an independent sample from some Dirichlet distribution parameterized by $\vec{\alpha}$. 
    \item Use an iterative method \cite{sklar2014dirichlet,minka2000dirichlet} to infer the parameters of the Dirchlet distribution from our sample vectors, as seen in Line 5 of Algorithm \ref{algorithm-ldm-entropy}. 
    \item Compute the entropy of the Dirichlet distribution we inferred. 
\end{enumerate}

The pseudocode for constructing an LDM of $M$ and estimating its complexity by measuring the entropy $\vec{\alpha}$ is below: \\


\noindent
Variable definitions:
\begin{align*}
        M &= \text{input classification model} \\ 
        K &= \text{number of datasets (columns) for LDM} \\
        D &= \text{input dataset}
\end{align*}

\begin{algorithm}
\caption{LDM Entropy Calculation Algorithm}
\label{algorithm-ldm-entropy}
\small
\begin{algorithmic}[1]
\For{$i = 1,\ldots,K$}
\State $f_i$, $H$ $\leftarrow$ splitIntoTrainHoldout(D)
\State $P_{f_i}$ $\leftarrow$ getSimplexVector(M, $f_i$, $H$)
\EndFor
\State L $\leftarrow$ $[P_{f_1}, P_{f_2},...,P_{f_K}]$
\State dirichletAlphas $\leftarrow$ findDirichlet(L)
\State entropy $\leftarrow$ computeEntropy(dirichletAlphas)
\end{algorithmic}
\end{algorithm}

We claim that this method can be used to measure model complexity. To give an intuitive sense of why, we first look at the labeling distribution matrix itself. If our model $M$ was trained on dataset $D_i$, then the output is a row vector $P_{\Omega}(D_i)$, indicating a probability distribution placed on $\Omega$ when $M$ was trained on $D_i$.

If we train on $K$ distinct datasets $D_1, D_2,..., D_K$, $M$ produces $K$ such column simplex vectors. Intuitively, if $M$ is a particularly expressive algorithm, then $M$ will be able to adapt well to different datasets, capturing the true relationships (and perhaps noise). As a result, we would expect more variance in the potential distributions over $\Omega$ that $M$ would be able to produce. On the other hand, if the simplex vectors in our algorithm's matrix are mostly very similar, then our algorithm $M$ does not have much capacity to adapt its parameters to the given dataset.

In general, if the LDM is ``compressible'' (containing redundant simplex vectors whose distributions are similar) then our algorithm is not very expressive. Thus, we consider the simplex vectors of our matrix themselves as samples from a Dirichlet distribution. If these vectors imply that the Dirichlet distribution covers most of its support, i.e., has a uniform distribution, then that means that our algorithm is as expressive as possible. This is because any vector in the support has the same probability assigned to it as any other vector, meaning that our algorithm is capable of producing anything in the support with equal likelihood. Therefore, the entropy of the inferred distribution is effectively a measure of the expressiveness of the algorithm that produced our labeling distribution matrix---a low entropy indicates a relatively less expressive algorithm, while high entropy implies a more expressive algorithm. 

\subsection{Results}

Using the methodology described in our experimental setup, we calculated the entropy of the labeling distribution matrices of a variety of machine learning models averaged over 20 runs. Every classification algorithm was trained on \texttt{scikit-learn}'s Iris Dataset \cite{fisher36lda}, which consists of 3 classes, 4 real-valued features for every example, and 150 examples total. Our Gaussian Process Classifier uses a default RBF kernel. Our Decision Tree uses a maximum depth of 5. Our Random Forest Classifier uses 10 estimators, a maximum of 1 feature, and a maximum depth of 5.

The entropies computed give us some insight as to a model's capacity and expressiveness. Every information resource on which we train a classification model is randomly generated, since we permute the labels of the original dataset when generating a column of an LDM. Thus, computing the entropy of $\vec{\alpha}$ parameterizing a Dirichlet distribution over an LDM gives us a measure of the diversity of distributions over labelings that the classification algorithm is capable of producing. The results are given in Table \ref{fig:model_entropy_table}.

\begin{table*}[ht]
\centering
\begin{tabular}{|l|c|}
\hline
\textbf{Model} & \textbf{Average Entropy of LDM} \\ \hline
Random Forest & -3100 \\ \hline
Gaussian Na\"ive Bayes & -1728 \\ \hline
Gaussian Process Classifier & -1244 \\ \hline
AdaBoost Classifier & -1264 \\ \hline
Quadratic Discriminant Analysis & -1128 \\ \hline
Decision Tree Classifier & -4908 \\ \hline
K-Nearest Neighbors (K = 10) & -1178 \\ \hline
K-Nearest Neighbors (K = 5) & -8429 \\ \hline
K-Nearest Neighbors (K = 3) & -18175 \\ \hline
K-Nearest Neighbors (K = 1) & -17227 \\ \hline
\end{tabular}
\caption{Average Entropy of LDM for a variety of classifiers}
\label{fig:model_entropy_table}
\end{table*}

\begin{figure}
\centering
\begin{subfigure}[b]{.45\linewidth}
\includegraphics[scale=0.38]{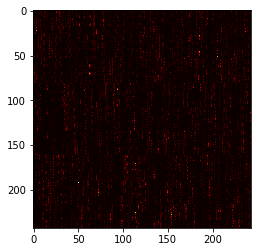}
\caption{Random Forest}\label{fig:randomForest}
\end{subfigure}
\begin{subfigure}[b]{.45\linewidth}
\includegraphics[scale=0.38]{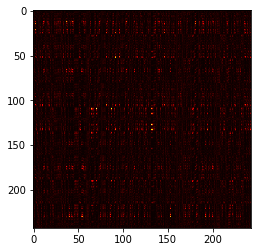}
\caption{Gaussian Naive Bayes}\label{fig:gaussianNaiveBayes}
\end{subfigure}

\begin{subfigure}[b]{.45\linewidth}
\includegraphics[scale=0.38]{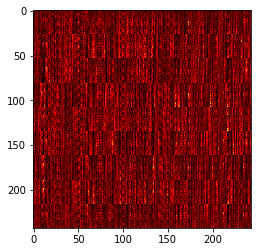}
\caption{Gaussian Process}\label{fig:gaussianProcess}
\end{subfigure}
\begin{subfigure}[b]{.45\linewidth}
\includegraphics[scale=0.38]{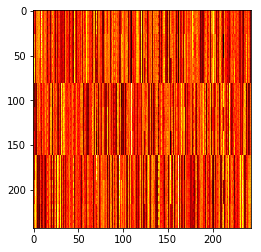}
\caption{AdaBoost}\label{fig:adaBoost}
\end{subfigure}

\begin{subfigure}[b]{.45\linewidth}
\includegraphics[scale=0.38]{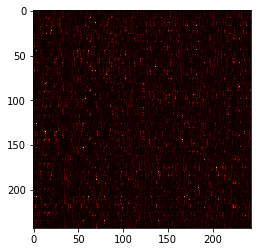}
\caption{QDA}\label{fig:qda}
\end{subfigure}
\begin{subfigure}[b]{.45\linewidth}
\includegraphics[scale=0.38]{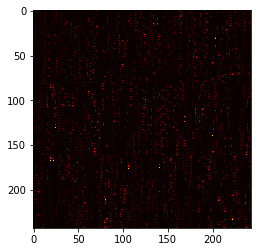}
\caption{KNN-3}\label{fig:knn3}
\end{subfigure}

\begin{subfigure}[b]{.45\linewidth}
\includegraphics[scale=0.38]{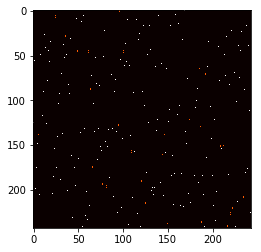}
\caption{Decision Tree}\label{fig:decisionTree}
\end{subfigure}

\caption{LDM Heatmaps for various classifiers}
\label{fig:ldmHeatMaps}
\end{figure}

Additionally, the LDM heatmaps shown in Figure \ref{fig:ldmHeatMaps} provides a visualization of the probability distributions of an LDM, 
with dark cells representing low probability, and brighter cells representing higher probability. Every index $i$ of the x-axis represents a particular labeling distribution $P_{f_i}$, while every index $j$ of the y-axis represents a unique labeling of the holdout set $H$, of which there are $C^{N'}$.

The average entropy of different classification models, as outlined in Figure \ref{fig:model_entropy_table}, coincides with what the average LDM looks like for the same model. Note, for example, that the highest entropy values are achieved by models like the Gaussian Process Classifier and AdaBoost Classifier. 

If we look at the Gaussian Process Classifier's LDM, as represented by the heatmap in Figure \ref{fig:gaussianProcess}, we see that the texture of the matrix is consistent (in color and pattern) over different information resources and labelings. Notice, it has an overall ``brightness'' that exceeds that of other classifiers, meaning that its models place more probability on different labelings than other models do.  This means that, despite being trained on different training datasets where the relationship between features and labels is broken or nonexistent, the Gaussian Process Classifier is still able to assign nonzero probability to most labelings of the holdout set. 

In similar vein, for the AdaBoost Classifier's LDM, as represented by the heatmap in Figure \ref{fig:adaBoost}, we see that the texture of the matrix is distinct from others, but still regular. It seems like there are three bands or sections regardless of the information resource that the model was trained on. In generating this matrix, we used a holdout set of size 5, where there were 3 classes. Since the ordering of the labelings along the y-axis are ordered lexicographically (with ``00000'' as the first labeling of the y-axis), the ridges of the matrix are potentially caused by the shift from labelings that start with a ``0'', to labelings that start with a ``1'', and finally to labelings that start with a ``2''. This is further verified by the fact that there are $3^4 = 81$ unique labelings of the holdout set that begin with a ``0'' and we can see that the first ridge of the heatmap occurs at exactly index $81$.

Conversely, the heatmap for the Decision Tree Classifier is dark overall, scattered with tiny spikes of light color. This means that its model confidently assigns most  probability mass to a single or few possible labelings, and no probability mass to other possible labelings. The Random Forest and KNN-3 classifiers have an appearance that is somewhere in between the extremes represented by the Gaussian Process Classifier and the Decision Tree Classifier. Comparing to the table, however, we see that the heatmap appearance doesn't fit neatly into the pattern of increasing entropy values for successive models.

The LDM attempts to capture two aspects simultaneously: the variation between output simplex vectors for models trained on different datasets, and the confidence (sparsity) within each individual simplex vector which determines how much probability mass is placed on the preferred labeling for a given dataset. In combining these two dimensions into a single number, the LDM estimation procedure loses some information, in that we do not know if a final high entropy value represents the averaging of many diverse high confidence vectors, or the averaging of less confident models, which produce homogeneous simplex vectors and lead to the same observed value. Flexible models should make confident predictions and be responsive to new data, leading to diverse output simplex vectors. Combining these features into a single number that measures capacity is the goal of the LDM process, of   which it is only partially successful.

\section{Shortcomings}

As notes in our Results section, the full LDM process seems to struggle in combining the two aspects of flexible models in an unambiguous way. Furthermore, we observe some values and trends which disagree with our traditional understanding of the relative flexibility of various methods. For example, KNN-1 should be the most prone to overfitting, having the greatest flexibility, yet its average entropy value is lower than that of KNN-10, which should be far more constrained and thus far less flexible. 

The problem could stem from one or more aspects of the procedure. Perhaps crucial information was lost as a result of averaging the values of the simplex vectors, as suggested in the previous section. In addition, by making algorithms output probabilities based on conditional independence of test instance labelings, this allows an algorithm like KNN-10 to place positive probability mass on many more individual test instances (likely having some nonzero number of neighbors with any chosen class label), whereas KNN-1 can only ever assign positive probability to the label of its single neighbor. Treating arbitrary simplex vectors as parameters for a Dirichlet model may also be problematic, since this modeling assumption was made for simplicity. 
 
Lastly, given the negative entropy values of the LDM process, it is difficult to understand these as positive capacity values, undermining the purpose for which the LDM was proposed. Negative entropy values can arise when using differential entropy, as when estimating the entropy of a continuous Dirichlet distribution. For the LDM-Dirichlet process to be used, one would still need a way of correlating entropy scores to storage capacity in bits.

\section{Future Work}
 
Given the aforementioned shortcomings of the LDM process and the continued need for methods of estimating algorithm capacity, other approaches should continue to be pursued. The question of how to estimate algorithm capacity is important, and failing to find a general solution to the question does not render the question any less important. 

One particularly promising idea, inspired by research in deep neural networks, is to use a form of autoencoder~\cite{doersch2016tutorial,olshausen1996emergence,lee2007efficient,bengio2014deep,bengio2013generalized,kingma2014stochastic} as applied to training data with labels that are independent of features. Generalization requires being able to predict labels given knowledge of the true relationship between features and labels. For a dataset with no relationship (in other words, with independence between features and labels), the only way an algorithm can reproduce the labels from the training dataset consistently is to memorize them, which it can do only in proportion to its capacity. Thus, for binary labels, the number of labels the algorithm can correctly retrieve in testing is the capacity (in bits) of how many labels it could memorize, plus some small number of luckily guessed labels (the number of which can be bound with high probability). 

\begin{table*}[ht]
\centering
\begin{tabular}{|l|c|c|}
\hline
\textbf{Model} & \textbf{Estimated Capacity} &  \textbf{95\% CI} \\ \hline
Random Forest & 144.79 &  [144.66, 144.92] \\ \hline
Gaussian Na\"ive Bayes & 60.79 & [60.53, 61.06]  \\ \hline
Gaussian Process Classifier & 75.00 &  [74.75, 75.26]  \\ \hline
AdaBoost Classifier & 92.58 &  [92.24, 92.91]\\ \hline
Quadratic Discriminant Analysis & 69.53 & [69.22, 69.83] \\ \hline
Decision Tree Classifier & 149.34 & [149.31, 149.37] \\ \hline
K-Nearest Neighbors (K = 10) & 73.26 & [72.91, 73.61] \\ \hline
K-Nearest Neighbors (K = 5) & 82.85 &  [82.50, 83.19] \\ \hline
K-Nearest Neighbors (K = 3) & 93.74 & [93.40, 94.08] \\ \hline
K-Nearest Neighbors (K = 1) & 149.34  & [149.31, 149.37]\\ \hline
\end{tabular}
\caption{Estimated capacity of a variety of classifiers using label recorders, with 95\% confidence intervals.}
\label{fig:model_capacity_LAE_table}
\end{table*}

A \textit{label recorder} takes randomly generated labels and independent training features comprising a training dataset, trains on that dataset, then tests on the same training set. The number of correctly reproduced labels will give a point estimate of the capacity, subject to random variation. Repeating this process and taking the average of the observed capacities will allow one to get an increasingly tighter estimate of the true algorithm capacity, arguably with fewer assumptions and steps than the LDM process. 

Table~\ref{fig:model_capacity_LAE_table} shows preliminary label recorder results for the models tested. Each method was tested on a set of 150 instances from the Iris dataset~\cite{fisher36lda}, with labels generated independently and uniformly at random. The point estimates were the average number of labels correctly recovered at test time, averaged over 1000 independent trials for each model. As can be seen from the table, we have unpruned Decision Trees and the Random Forest Classifier having the highest estimated capacity, while more bias-heavy models such as Quadratic Discriminant Analysis and Gaussian Na\"ive Bayes have less capacity. Furthermore, the estimated capacities for KNN as a function of the regularization parameter $K$ show decreasing capacity with increasing $K$, aligning better with our intuition than the LDM inferred entropies. Thus, label recorders present a promising avenue for estimating algorithm capacity.
Creating label recorders and using them to provide rigorous bounds on algorithm capacity is the subject of future work, which we hope will complement (if not supersede) the work presented here.

\section{\uppercase{Conclusions}}
\label{sec:conclusion}

\noindent In an attempt to estimate the capacity of algorithms, as a measure of the amount of data their models can store, we introduce the LDM. We tested the LDM procedure on several learning models and observed the average entropy values over a variety of datasets. Our results highlighted interesting behaviors of the algorithms tested, but were not as conclusive or consistent as initially hoped.

Although the LDM process falls short in several respects, the underlying idea of a Labeling Distribution Matrix could prove useful in developing other methods for measuring algorithm capacity. Perhaps other ways of averaging or combining data from the LDM could lead to better, more interpretable results. Recent theoretical work on \textit{entropic expressivity} may be one such application of the LDM \cite{bias-expressivity}. Thus, even though the LDM process is not a perfect capacity estimation method, the LDM object itself might still prove useful as a component of such a method. Lastly, the label recorders briefly presented here hold promise for serving as a method of estimating algorithm capacity, as shown through our preliminary results.

\vfill


\bibliographystyle{apalike}
{\small
\bibliography{main}}

%

\end{document}